Review

# A Survey of Methods for Managing the Classification and Solution of Data Imbalance Problem

**[1]Khan Md. Hasib, [2]Md. Sadiq Iqbal, [3]Faisal Muhammad Shah, [4]Jubayer Al Mahmud, [5]Mahmudul Hasan Popel, [6]Md. Imran Hossain Showrov, [7]Shakil Ahmed and [8]Obaidur Rahman**

[1,3,5]*Department of Computer Science and Engineering, Ahsanullah University of Science and Technology, Bangladesh*
[2,7,8]*Department of Computer Science and Engineering, Bangladesh University, Bangladesh*
[4]*Department of Computer Science and Engineering, University of Dhaka, Bangladesh*
[6]*Institute of Computer Science, Bangladesh Atomic Energy Commission, Bangladesh*

Corresponding Author:
Md. Imran Hossain Showrov
Institute of Computer Science,
Bangladesh Atomic Energy
Commission, Bangladesh
Email: showrov@baec.gov.bd

**Abstract:** The problem of class imbalance is extensive for focusing on numerous applications in the real world. In such a situation, nearly all of the examples are labeled as one class called majority class, while far fewer examples are labeled as the other class usually, the more important class is called minority. Over the last few years, several types of research have been carried out on the issue of class imbalance, including data sampling, cost-sensitive analysis, Genetic Programming based models, bagging, boosting, etc. Nevertheless, in this survey paper, we enlisted the 24 related studies in the years 2003, 2008, 2010, 2012 and 2014 to 2019, focusing on the architecture of single, hybrid and ensemble method design to understand the current status of improving classification output in machine learning techniques to fix problems with class imbalances. This survey paper also includes a statistical analysis of the classification algorithms under various methods and several other experimental conditions, as well as datasets used in different research papers.

## Introduction

The data is the most valuable asset and one of the best possible sources for any research and development. Data can be viewed as one of the significant components of educational and business strategy decisions. Therefore, the data for research and development for any major decision should be balanced and accurate. Balanced data is one of the major concerns nowadays. Data Imbalance problems impede the performance of the classification algorithm (Singh and Purohit, 2015). However, the efficiency of predictive models is significantly impacted when the data set in the real world is highly imbalanced (Amin *et al.*, 2016). So, the data are used for strategic decisions and research should be balanced. When class distribution in the dataset is not uniform, the data are called imbalance (Haykin, 1999). In such cases, there are only a limited number of instances represented in the least one known as minority class and the remainder of the dataset consisting of other classes is known as majority class. Recent work has shown that output an unequal distribution of class examples in the learning process can skew efficiency.

This means the class provides minimal specificity on the minority class while it offers great accuracy in the majority class. Class disparity in the datasets can drastically skew the performance of classifiers in a majority-minority class imbalance problem, introducing a prediction bias for the majority class (Leevy *et al.*, 2018). Though high imbalance affects the output significantly, some of the small imbalances metrics were beneficial (Koziarski *et al.*, 2018).

Researchers divided the data imbalance problem into two major categories: Multiclass data imbalance (Bhowan *et al.*, 2011). In a multiclass dataset, there are more than two classes and just two classes in a binary dataset. There are plenty of attempts to fix the issue of binary class imbalance, but the different types of problems relevant to the problem of multiclass imbalance are not yet solved (Rout *et al.*, 2018). Learning from imbalanced data is studied extensively in standard classification and also in multilevel classification in recent times (Charte *et al.*, 2019). There are two major approaches (external and internal) used to build methods to solve the problem of data imbalances (Eggermont *et al.*, 2004). Bagging and boosting are the most common external approaches for handling the data imbalance problem sampling. Researchers have proposed only a few approaches within their internal approach. Among them, Genetic Programming (GP) (Haykin, 1999) and the Extreme Learning Machine (ELM) (Yu *et al.*, 2016) approaches are the most popular techniques to solve the data imbalance problem.

Over the years, many scholars and researchers have done some successful work on the improvement of the imbalance dataset. This study analyzed all of the relevant research in the data collection of imbalances over the last separate ten years. In total 24 articles are enlisted in this survey paper from 2003, 2008, 2010, 2012 and 2014 to 2019. This study contained the proposed method of classification techniques with learning algorithms. Statistical comparison was provided for viewing class if the technique and the algorithm chosen were used along with the selection steps of the function. The paper is set out as follows: An overview of the research subject is given in section 2, which describes a variety of imbalanced dataset techniques. Section 3 offers a statistical analysis of the papers where various year-wise approaches are discussed. Section 4 includes discussion and conclusion alongside indicating some issues for future imbalance dataset research using approaches to machine learning.

## Research Paper Overview

A number of researches are going on in the field of data mining and machine learning such as keyword extraction (Showrov and Sobhan, 2019), summarization (Abulaish *et al.*, 2018; Showrov *et al.*, 2019a), breast cancer detection (Showrov *et al.*, 2019b) and so on. The most challenging problem nowadays in this field is a class imbalance. Several scholars have suggested various types of approaches for dealing with problems with class imbalances. Methods of data level, methods of leveling algorithms and methods of the ensemble are categorized methods. Hybrid methods are another group type for dealing with the problem of class imbalances.

### Data Level Methods

This approach is geared towards matching the class distributions. The class distribution are being balanced using the sampling methods by resizing the training datasets. The sampling methods can be categorized into techniques for under-sampling and over-sampling.

### Over-Sampling Technique

Oversampling is the method of either randomly increasing the number of instances in the minority class to increase the disparity ratio such that the corresponding classification algorithms can be employed for the classification of the data. The benefit of this technique is that any necessary information is not missed from the dataset and the primary dataset can be preserved even though new data is appended to it for balancing the data (Kaur and Gosain, 2018).

### SMOTE

Synthetic Minority Over-sampling Technique (SMOTE) is a process to increase the data into the minority class by generating new synthetic data using the existing data (Junsomboon and Phienthrakul, 2017). It is capable of producing patterns that follow a distribution similar to the true on (Elreedy and Atiya, 2019). However, other methods, such as fuzzy or locally linear embedding (Verbiest *et al.*, 2012), continue to improve SMOTE.

### Cluster-Based Over-Sampling (CBO)

This approach contains clustering the training data of each class separately that is achieved using the k-means method. Thereafter, random over-sampling is carried out on all clusters (Popel *et al.*, 2018).

### ADASYN

Adaptive synthetic sampling method improves learning on the distributions of data in two ways: (1) Reduction the biasness which is introduced by the class imbalance and (2) adaptively is shifting the boundary of classification decisions towards the difficult examples (He *et al.*, 2008).

### Under-Sampling Technique

In the under-sampling method, the working area in the dataset is a majority class where either randomly or by using some technique to balance the classes are extracted from the majority class. The under-sampling method is used to boost the imbalance ratio on unbalanced data and classes are then classified using conventional classification algorithms (Kaur and Gosain, 2018). However, it does have the benefit of reducing the time required to train the models because the size of the training data set is reduced (Seiffert *et al.*, 2009).

### Random Under-Sampling (RUS)

Random Under-Sampling (RUS) is an under-sampling method that excludes the majority-class instances randomly to balance the class distribution (Popel *et al.*, 2018). It is a technique that removes examples randomly from the class of majority. Given its simplicity, RUS was shown to perform very well. Simplicity, speed and efficiency are the reasons for the introduction of RUS into the boosting process (Seiffert *et al.*, 2009).

### Tomek Link (T-Link)

T-Link is a technique of under-sampling stated by Tomek. This is seen as improving the Nearest-Neighbor

Regulation (NNR). The T-Link technique can be used as a directed under-sampling method where the majority-class observations are deleted (Rahman *et al.*, 2011).

### Algorithm Level Methods

Algorithm level approaches concentrate on improving the ability of current classifier algorithms for learning from minority classes, which are often called internal approaches. For example, adjustment of the estimation of probability or modification of cost per class may be favorable to the minority class.

### Support Vector Machine (SVM)

Support Vector Machine was introduced in the mid-1990s (Rahman *et al.*, 2011). This technique discriminates over input spaces in a finite area. It is necessary for classifications to be obtained by learning from the training sample (Durgesh and Lekha, 2010). Traditional SVM classification methods use as input training data consisting of a mix of data classified by two groups (Catania *et al.*, 2012).

### K-Nearest Neighbor (KNN)

A number of distance measuring methods are being adopted in K-nearest neighbors. In the training results, KNN finds *k* number of nearest samples and then allows the class label used frequently within the estimated training samples based on the test sample. The K-nearest neighbor is known for being the simplest and most non-parametric sample classification (Friedl and Brodley, 1997). KNN can be mentioned as a learner based on the instance (Bishop, 1995).

### Naïve Bayes

To simplify the relationship, Naïve Bayes often produces strong classification outcomes. Although a lot of classification missions, only one scan of the training data is required (Mitchell, 1997). Based on the given class label, Naïve Bayes estimates that. The attributes are independent of conditions and therefore investigate to determine the class conditional probability (Kotu and Deshpande, 2018).

### Decision Tree

A Decision tree classifies an unknown test instance by way of a series of decisions. Decision tree classifiers are widely used in many different ways, in particular for their high adaptability to complex classification problems (Friedl and Brodley, 1997). The decision tree is simpler and easier to enforce, so as a single classifier it is renowned (Farid *et al.*, 2013).

### Ensemble Methods

Ensemble approaches involve the synthesis of various methods. Ensembles based on bagging and boosting are commonly used and are efficient solutions for the class issue with imbalances. Breiman (1996) presented the idea of aggregating bootstraps to create ensembles.

### Bagging

Bagging (Bootstrapped Aggregating) is a way to boost the classification algorithms results (Machová *et al.*, 2006). Bagging utilizes and integrates multiple self-employed learners using an averaging technique. Reducing variation and bias (Sanabila and Jatmiko, 2018) works fine.

### Boosting

Schapire (1990) launched Boosting. Schapire has shown that a low learner (slightly better than guessing randomly) can transform into a powerful learner. AdaBoost is the family's most influential algorithm. Boosting needs bootstrapping too. There is yet some other difference here, though. Unlike bagging, each sample of data boosts weights. It means that some samples will be run more frequently than others (Breiman, 1996).

### Hybrid Methods

The hybrid approaches include both data sampling and algorithm boosting. While many data sampling techniques are specifically designed to address the problem of class imbalances, hybrid methods can improve the performance of any weak classifier (regardless of whether the data is unbalanced) (Seiffert *et al.*, 2009).

### SMOTEBoost

SMOTEBoost produces synthetic examples of the rare or minority class, thereby implicitly adjusting the weights of updates and compensating for distorted distributions. This construct focuses on the sampled minority class examples for each boosting iteration and creates new examples (Chawla *et al.*, 2003).

### RUSBoost

Random Under-Sampling Based Boosting (RUSBoost) Method presents a simple, quicker and less complex method for learning from imbalanced data. It proposed a detailed empirical study comparing the performance of several strategies for improving the efficiency of classification when data is imbalanced (Seiffert *et al.*, 2009).

### LIUBoost

Locality Informed Underboosting (LIUBoost) approach combines sampling technique with cost-sensitive learning. Under-sampling, it uses data sets in every boosting iteration such as RUSBoost, while incorporating a cost term for each instance based on their hardness into the weight update formula minimizing the information loss introduced by under-sampling (Ahmed *et al.*, 2019).

*RHSBoost*

For further enhancement of classification accuracy, Random Hybrid Sampling Boosting (RHSBoost) uses both under-sampling and ROSE sampling in the boosting algorithm. Under a boosting scheme (Gong and Kim, 2017), the classification rule uses random under-sampling and ROSE sampling.

*HUSBoost*

Hybrid Under-Sampling Based Boosting (HUSBoost) approach proposes managing imbalanced data that requires three basic steps-data cleaning, data balance and classification. The goal of these methods is to optimize the overall accuracy while these algorithms neglect the minority class most of the time (Popel *et al.*, 2018).

## Statistical Comparison of Related Work

### Distribution of Papers by Year of Publications

This survey contains 24 research articles within the period from 2003, 2008, 2010, 2012 and 2014 to 2019. It addressed 3 papers for 2014 and 2019, each year. From the year 2017, researchers studied the largest number of papers. For that year, the number of papers is 6. Figure 1 shows the ratio of paper distribution by the released year.

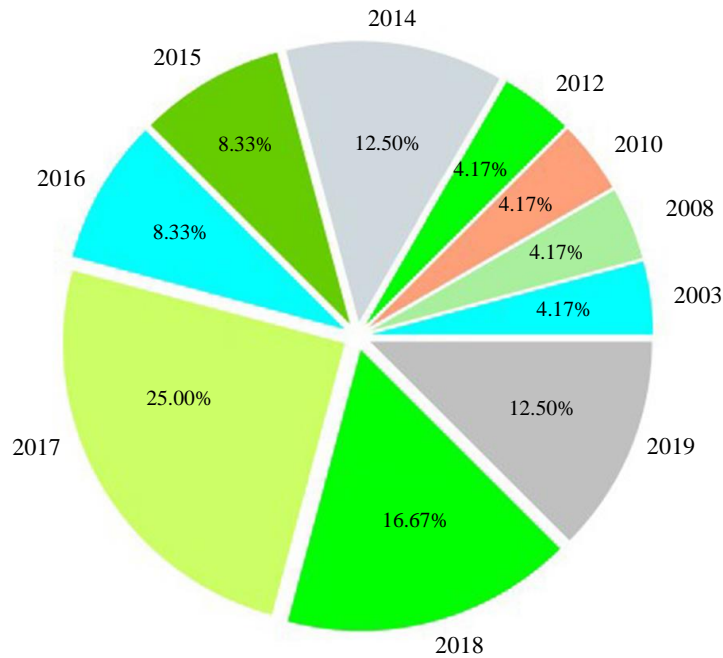

**Fig. 1:** Distribution of papers based on years

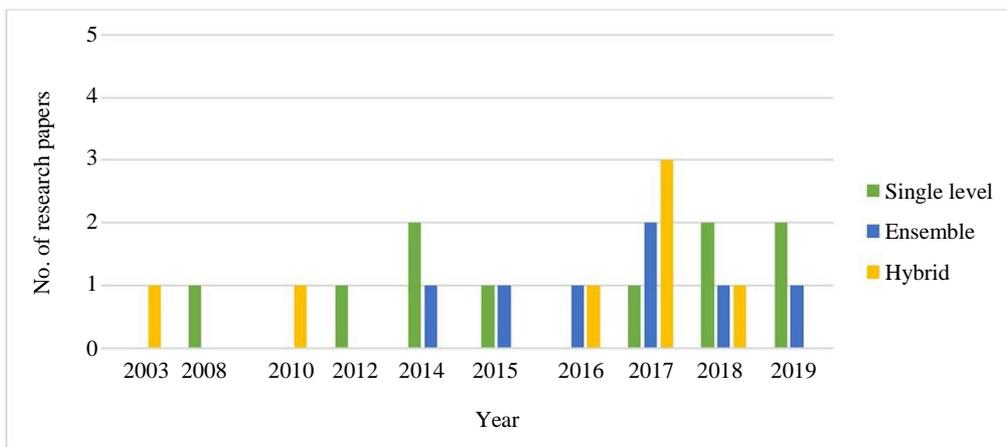

**Fig. 2:** Distribution of papers based on years

1549

**Table 1:** Total number of research papers for the standards of method design

| Method design type | No. of research paper | References |
|---|---|---|
| Single level (both data level algorithm level) | 10 | Hu *et al.* (2015; Jedrzejowicz *et al.*, 2018; Maldonado *et al.*, 2014; Fernández *et al.*, 2017; Ebenuwa *et al.*, 2019; Charte *et al.*, 2019; Koziarski *et al.*, 2018; Barua *et al.*, 2012; Verbiest *et al.*, 2012; He *et al.*, 2008) |
| Ensemble | 7 | Sun *et al.* (2015; Yijing *et al.*, 2016; Wei *et al.*, 2017; Yu and Ni, 2014; Lango and Stefanowski, 2017; Wang, 2019; Sanabila and Jatmiko, 2018) |
| Hybrid | 7 | Ahmed *et al.* (2019; Junsomboon and Phienthrakul, 2017; Elhassan and Aljurf, 2016; Gong and Kim, 2017; Popel *et al.*, 2018; Seiffert *et al.*, 2009; Chawla *et al.*, 2003) |

*Method Design*

The unbalanced set of data can be categorized into various categories, namely single, ensemble and hybrid. Type of single-level design includes 10 papers, the type of ensemble design covers 7 papers and finally, the type of hybrid design contains 7 papers. The enlisted 24 papers are given at a glance in the following Table 1. Figure 2 represents the number of research papers based on single, ensemble and hybrid methods used in each particular year.

*Single Level Method*

The number of research papers utilizing different types of sampling classifiers and traditional machine learning algorithms in the single-level method. Mostly, the single level process survey uses Synthetic Minority Over-sampling Technique (SMOTE), Decision Tree (DT), Support Vector Machine (SVM), K-Nearest Neighbor (KNN), Edited Nearest Neighbor (ENN) and Naïve Bayes.

Table 2 depicts the year-wise distribution of the Single level method regarding advantage or limitation and citation.

*Ensemble Method*

Multiple algorithms are combined in this method. Table 3 reflects the year-wise distribution of the ensemble method concerning the benefit or disadvantage that we are analyzing here. Adaboost, ROS, RUS, etc. are different types of algorithms used in ensemble methods. Table 3 also lists the method proposed for each paper and the citation of each article.

*Hybrid Method*

Table 4 shows the year-wise distribution of the hybrid method concerning the results and citations of each paper, the hybrid method is used to solve the imbalanced data set in the mainstream study due to the recent output accuracy. Statistics show the largest number of publications on the hybrid approach in 2017. The table also displays in each article the algorithm used and their success in solving the problem of an imbalanced dataset.

*Used Dataset in Researches*

Datasets are assigned for default tasks such as classification, clustering, prediction results, etc. This survey paper analyses Dataset is for classification purposes. Table 5 displays the distribution of randomly used datasets year by year.

KDD cup 1999 dataset (Chawla *et al.*, 2003) has multivariate data types and contains 4,000,000 instances and 42 attributes. Abalone dataset (He *et al.*, 2008) contains 4,177 instances with 8 attributes, attributes types are categorical. Glass dataset (Hu *et al.*, 2015) with multivariate data types and 214 instances. *E coli* 2 dataset (Elhassan and Aljurf, 2016) carries 363 instances and 7 attributes. Ionosphere dataset (He *et al.*, 2008) contains 351 instances with 34 attributes, attributes types are integer and real. Wisconsin breast cancer dataset (Ebenuwa *et al.*, 2019) has multivariate data types, all 10 instances are integer types and it has 699 instances. Yeast dataset (Hu *et al.*, 2015) have 8 real attributes with 1,484 instances. Various kinds of datasets from the Keel dataset repository (Verbiest *et al.*, 2012; Ahmed *et al.*, 2019; Gong and Kim, 2017; Jedrzejowicz *et al.*, 2018; Fernández *et al.*, 2017; Wang, 2019) are mostly used in handling imbalanced datasets. Liver-Disorders-Bupa (Ebenuwa *et al.*, 2019) contains 345 instances with 7 attributes where attribute types are Categorical, integer and real.

The study also shows that few private or non-public datasets are used over the time frame. Although the study briefly highlights the UCI machine learning repository datasets being considered as standard datasets for handling and solving imbalanced data. Some medical datasets such as Ovarian I and Ovarian II (Yu and Ni, 2014) datasets are used as well as Breast Cancer, ILPD, Pima Indians, Fertility, Haberman form medical dataset (Junsomboon and Phienthrakul, 2017) are also used. On the other hand, customer behavior of banking transaction Data (Sanabila and Jatmiko, 2018) is used to get the accuracy.

*Discussion of Surveyed Works*

The surveyed works and their specific data sets were summarized in Table 2 to 5 to provide a high-level overview and better compare the existing approach for various learning models in class imbalance. Table 2 points out on single-level methods which involve both data level and algorithm level approaches where different kinds of the method are used to balance the imbalanced datasets.

**Table 2:** Year-wise distribution of single level method regarding advantage/limitation and citation

| Year | Research paper title | Reference | Algorithm used | Proposed method | Advantage/limitation | Citation |
|------|---------------------|-----------|----------------|-----------------|---------------------|----------|
| 2008 | ADASYN: Adaptive synthetic sampling approach for imbalanced learning | He *et al.* (2008) | • DT | ADASYN | Advantage:<br>• Can be used to multiple- class imbalanced learning<br>• Can be modified<br>Limitation:<br>• Ensemble learning is more effective then single learning | 843 |
| 2012 | Improving SMOTE with fuzzy rough prototype selection to detect noise in imbalanced classification data | Verbiest *et al.* (2012) | • KNN | FRIPS | Advantage:<br>• FRIPS can handle noise<br>• FRIPS improves SMOTE<br>Limitation:<br>• Not better than SMOTE-TL | 12 |
| 2014 | Feature selection for high-dimensional class-imbalanced data sets using Support vector machines | Maldonado *et al.* (2014) | • SVM | BFE-SVM, HO-BFE | Advantage:<br>• Good results on highly imbalanced data sets<br>• Minimize the number of errors in the minority class<br>Limitation:<br>• Kernel-based methods expose very unstable performance<br>• HO-BFE version running time is about five times higher | 107 |
| 2012 | MWMOTE-Majority Weighted minority oversampling technique for imbalanced data set learning | Barua *et al.* (2012) | • NN<br>• DT<br>• ENN<br>• KNN | MWMOTE | Advantage:<br>• Boosting improves the recall performance<br>• Selects the hard-to-learn minority class<br>Limitations:<br>• Use data sets with continuous features only | 261 |
| 2015 | An improved algorithm for Imbalanced data and small sample size classification | Hu *et al.* (2015) | • KNN<br>• SVM | WRO | Advantage:<br>• Can enlarge the decision regions<br>• Improve the prediction of the minority class<br>Limitations:<br>• Too many parameters<br>• No guidelines for deciding relative ratios of cost factors | 9 |
| 2017 | An insight into imbalanced big data classification: Outcomes and challenges | Fernández *et al.* (2017) | • KNN<br>• DT | MapReduce | Advantage:<br>• Handle multi-class imbalance<br>Limitations:<br>• Computationally expensive<br>• Lack of sufficient data in the training partitions | 71 |
| 2018 | Imbalanced data classification using MapReduce and relief | Jedrzejowicz *et al.* (2018) | • Naive Bayes<br>• LR<br>• DT | MapReduce Relief | Advantage:<br>• Binary classification<br>• Does not change the quality of classification<br>Limitations:<br>• Longer processing time | 91 |
| 2018 | Network-Based classification of histopathological images affected by data Imbalance | Koziarski *et al.* (2018) | • SMOTE | CNN | Advantage:<br>• Suitable for low data imbalance level<br>Limitations:<br>• Significantly underperformed<br>• Not suitable for medium and high data imbalance level | 6 |
| 2019 | Variance ranking attributes selection techniques for binary classification problem in imbalance data | Ebenuwa *et al.* (2019) | • LR<br>• SVM<br>• DT | ROS | Advantage:<br>• Higher accuracy<br>• Achieving predictions with fewer attributes<br>Limitations:<br>• Variable must be numeric<br>• Variable must not be categorical | 12 |
| 2019 | Tackling multilabel imbalance through label decoupling and data resampling hybridization | Charte *et al.* (2019) | • BR<br>• LP<br>• KNN | REMEDIAL | Advantage:<br>• Improves the efficiency of oversampling<br>• Improve the training: BR and LP<br>Limitations:<br>• Not able to balance a high imbalanced labels distribution<br>• Produce new instances in the minority labels only | 8 |

**Table 3:** Year-wise distribution of ensemble method regrading advantages/limitations and citation

| Year | Research paper title | Reference | Algorithm used | Proposed method | Advantage/limitation | Citation |
|---|---|---|---|---|---|---|
| | An improved ensemble learning method for classifying high-dimensional and imbalanced biomedicine data | Yu and Ni (2014) | • SVM | asBagging FSS | Advantage:<br>• Greatly enhance the diversity<br>• Improve the balance level<br>Limitation:<br>• Not suitable for low-dimensional data | 20 |
| 2015 | A novel ensemble method for classifying imbalanced data | Sun *et al.* (2015) | • Naive Bayes<br>• DT<br>• KNN | ClusterBal SplitBal | Advantage:<br>• Significantly better than Orig method<br>• Able to handle the binary-class<br>Limitations:<br>• G-Means and F-Measure performance were not selected | 137 |
| 2016 | Adapted ensemble classification algorithm based on multiple classifier system and feature selection for classifying multi-class imbalanced data | Yijing *et al.* (2016) | • SVM<br>• DT<br>• ANN | AMCS | Advantage:<br>• Choose best route for different types of data<br>Limitations:<br>• Did not employ cost sensitive learning | 53 |
| 2017 | An ensemble model for diabetes diagnosis in large-scale and imbalanced datasets | Wei *et al.* (2017) | • LR<br>• CART<br>• LSVC<br>• Adaboost<br>• RF<br>• XGB | xEnsemble | Advantage:<br>• Reduce the variance by under-sampling<br>• Excellent performance than Easy Ensemble<br>Limitations:<br>• Discard much potentially useful data<br>• Weak classifier and running serially | 14 |
| 2017 | Multi-class and feature selection extensions of Roughly balanced bagging for imbalanced data | Lango and Stefanowski (2017) | • Naive Bayes<br>• SVM<br>• LR<br>• DT | RBBag+ RSM | Advantage:<br>• Better on selected attributed<br>• Does not influence the final performance<br>Limitations:<br>• Component classifiers characterized by quite low diversity<br>• Semi-supervised learning | 23 |
| 2018 | Ensemble learning on large scale financial imbalanced data | Sanabila and Jatmiko (2018) | • SMOTE<br>• ENN | Resampling | Advantage:<br>• Help to minimize the bias to the majority class<br>• Robust result and less misclassification<br>Limitations:<br>• Resampling hinders the performance of other learning methods | 4 |
| 2019 | An ensemble learning imbalanced data classification method based on sample combination optimization | Wang (2019) | • SMOTE<br>• LR<br>• ROS<br>• RUS | GABagging | Advantage:<br>• Remove overlapping regions of different categories<br>• Good prediction effect<br>Limitations:<br>• Higher time complexity<br>• Easy loss<br>• Increasing samples | 11 |

ADASYN (He *et al.*, 2008) and FRIPS (Verbiest *et al.*, 2012) methods are used to handle multiple class imbalance learning and to control noise to classify imbalanced data using the like of traditional algorithm DT and KNN. Single-level approaches have been investigated in 10 studies and can be further divided into new loss functions, cost-sensitive approaches, performance thresholds. Ensemble methods involve 7 studies that are depicted in Table 3. Resampling method (Sanabila and Jatmiko, 2018) were combined SMOTE and ENN to minimize the bias of majority class and less misclassification in the way of classifying and also AMCS (Yijing *et al.*, 2016) method used Naïve Bayes, DT and KNN to choose the best route for different types of data for classifying multiclass imbalanced data. Multiple authors (Popel *et al.*, 2018; Ahmed *et al.*, 2019) suggested that the use of a machine learning model with DT to address the class imbalance in the field of hybrid methods which denotes in Table 4. A combination of sampling, cost-sensitive method makes the hybrid method better sometimes to give better accuracy than the others in case of balancing data.

**Table 4:** Year-wise distribution of hybrid method regarding advantage/limitation and citation

| Year | Research paper title | Reference | Algorithm used | Proposed method | Advantage/Limitation | Citation |
|---|---|---|---|---|---|---|
| 2003 | SMOTEBoost: Improving prediction of the minority class in boosting. | Chawla *et al.* (2003) | • RIPPER | SMOTEBoost | Advantage: <br> • Minimal accuracy degradation <br> • Create new synthetic examples from the minority class <br> Limitation: <br> • Not effective for skewed class <br> • Chose the smallest class as the minority class and collapsed the remaining classes into one class | 828 |
| 2009 | RUSBoost: A Hybrid approach to alleviating class Imbalance | Seiffert *et al.* (2009) | • C4.5 <br> • DT <br> • Naive Bayes | RUSBoost | Advantage: <br> • Can learning from skewed training data <br> • Better classification performance than AdaBoost <br> Limitations: <br> • Increased model training time on larger training data sets <br> • Loss of information | 669 |
| 2016 | Classification of Imbalance data using Tomek Link (T-Link) combined with Random Under-Sampling (RUS) as a Data Reduction Method | Elhassan and Aljurf (2016) | • LR <br> • SVM <br> • ANN <br> • RF | T-Link | Advantage: <br> • Methods of data reduction <br> • Superior performance <br> Limitations: <br> • Majority class removed <br> • Information loss | 13 |
| 2019 | LIUBoost: Locality Informed Under-Boosting for imbalanced data classification | Ahmed *et al.* (2019) | • KNN <br> • DT | LIUBoost | Advantage: <br> • Cost-efficient <br> • Minimize combined error <br> • Minimizing information loss <br> Limitations: <br> • Alpha term needs to be update <br> • Suffers overfitting <br> • Increased runtime | 4 |
| 2017 | Combining over-sampling and under-sampling techniques for imbalance dataset | Junsomboon and Phienthrakul (2017) | • KNN <br> • Naive Bayes | NCL+ SMOTE | Advantage: <br> • Help to increase correct to classify <br> • Good prediction in minority class <br> • Better performance on large dataset <br> Limitations: <br> • Few numbers of sample data <br> • Possible to ignore serious data <br> • Information loss | 12 |
| 2017 | RHSBoost: Improving classification performance in imbalance data | Gong and Kim (2017) | • NN | RHSBoost | Advantage: <br> • Improve classification accuracy <br> • Relatively high performance <br> • Less vulnerable <br> Limitations: <br> • Slightly weaker for large iterations | 36 |
| 2018 | A Hybrid Under-Sampling Method (HUSBoost) to classify imbalanced data | Popel *et al.* (2018) | • RF <br> • SVM <br> • DT | HUSBoost | Advantage: <br> • Clean noisy and overlapping data <br> • Consider several balanced subset <br> Limitations <br> • Arithmetic mean taken <br> • Higher running time | 2 |

**Table 5:** Year wise distribution of randomly used dataset

| Dataset | 2003 | 2008 | 2015 | 2016 | 2018 | 2019 | Total |
|---|---|---|---|---|---|---|---|
| KDD cup 1999 | 1 | | | | | | 1 |
| Glass | | | 1 | | | | 1 |
| Ecoli2 | | | | 1 | | | 1 |
| Abalone | | 1 | | | | | 1 |
| Wisconsin breast cancer | | | | | 1 | | 1 |
| Yeast | | | 1 | | | | 1 |
| Ionosphere | | 1 | | | | | 1 |
| Liver-disorder-bupa | | | 1 | | | | 1 |

About half of the researchers used just various imbalance to test their machine learning approaches to resolve the class disparity. The writers are not at fault for this, as most of them have been concentrating on addressing a particular problem or a benchmark assignment. However, more rigorous experiments testing these approaches across a broader variety of data sets, with differing degrees of class inequality and difficulty, would help explain their strengths and limitations. Also, only one-third of the experiments show the number of rounds or repetitions done on each trial. In other words, the rest of it the groups either did not conduct several runs or neglected to provide the specifics and reported the most desirable findings. For their safety, training a machine learning model on a broad data set will take days or even weeks, making it difficult to perform many rounds of experiments. This opens a range of opportunities for future study, as applying different machine learning approaches to several data sets with replication can improve confidence in outcomes and help direct future practitioners in model selection.

The methods that have been discussed explicitly cannot currently be comparable, since they are tested in separate data sets with differing class imbalances and results with contradictory performance assessments are published. Besides, some studies report inconsistent findings, further indicating that performance is highly dependent on problem complexity, class representation and recorded performance metrics. Overall, there is a lack of evidence that distinguishes any hybrid, ensemble machine learning method as superior for learning from class imbalanced results and additional experiments are needed before such conclusions can be drawn. Class imbalance is not limited to low or high imbalance data and further analysis needs to be performed to test the application of these machine learning class imbalance approaches in other domains.

## Conclusion and Future Works

Data imbalances are a common concern. It has been a long time since the attention of the researchers. Uses of various imbalance dataset classifier techniques are an emerging data science study. Existing classifier performance on an imbalanced dataset is not expected without a balancing dataset. So, we have to reprocess the data to make a better accuracy of results. In this survey, the application of using different classifiers has been established for the classification of imbalances in results. This survey paper has provided a clear comparison of these papers and a fair viewpoint in this area, but this study cannot say an in-depth analysis of those papers. The following points may be useful in future research:

- System performance is a key factor. In the training phase, the removal of redundant and irrelevant features increases system performance

- In the classification techniques, the consideration feature selection step will play a vital role in the future
- For performance measurement, uses of hybrid or ensemble classifiers are more feasible instead of a single classifier

There are some areas for future work that are evident. Applying the newly developed methods to a wider range of data sets and class imbalance levels, comparing outcomes with several complementary performance indicators and reporting statistical evidence can help to define the optimal deep learning approaches as well as traditional machine learning methods for future class imbalance applications. Experimenting with new hybrid and cluster-based machine learning approaches along with deep learning approaches to fix class imbalances in the sense of big data and class rareness may prove useful for the future of big data analytics.

## Acknowledgment

We would like to thank Google and UCI Machine for providing the dataset and necessary information for this this research.

## Author's Contributions

**Khan Md. Hasib:** He contributed to all the sections of the paper. He worked on the design and implementation. He participated in correcting the paper and responding to all reviewers' comments.

**Md. Sadiq Iqbal:** He was involved in data gathering, experimentation and organization of the paper.

**Faisal Muhammad Shah:** He contributed to all the sections of the paper. He coordinated the data gathering, design and execution of all experiments. He also participated in correcting the paper and responding to all reviewers' comments.

**Jubayer Al Mahmud:** He contributed to all the sections of the paper. He coordinated the data gathering, design and execution of all experiments. He also participated in correcting the paper and responding to all reviewers' comments.

**Mahmudul Hasan Popel:** He contributed to all the sections of the paper. He coordinated the data gathering, design and execution of all experiments. He also participated in correcting the paper and responding to all reviewers' comments.

**Md. Imran Hossain Showrov:** He contributed to all the sections of the paper. He conceived the idea and worked on the implementation. He was involved in data gathering, experimentation and organization of the paper. He participated in correcting the paper and responding to all reviewers' comments.

**Shakil Ahmed:** He contributed to all the sections of the paper. He coordinated the data gathering,

design and execution of all experiments. He also participated in correcting the paper and responding to all reviewers' comments.

**Obaidur Rahman:** He was involved in data gathering, experimentation and organization of the paper.

## Ethics

This article is original and contains unpublished material. The corresponding author confirms that all of the other authors have read and approved the manuscript and no ethical issues involved.

## References

Abulaish, M., Showrov, M. I. H., & Fazil, M. (2018, November). A layered approach for summarization and context learning from microblogging data. In Proceedings of the 20th International Conference on Information Integration and Web-based Applications & Services (pp. 70-78).

Ahmed, S., Rayhan, F., Mahbub, A., Jani, M. R., Shatabda, S., & Farid, D. M. (2019). Liuboost: Locality informed under-boosting for imbalanced data classification. In Emerging Technologies in Data Mining and Information Security (pp. 133-144). Springer, Singapore.

Amin, A., Anwar, S., Adnan, A., Nawaz, M., Howard, N., Qadir, J., ... & Hussain, A. (2016). Comparing oversampling techniques to handle the class imbalance problem: A customer churn prediction case study. IEEE Access, 4, 7940-7957.

Barua, S., Islam, M. M., Yao, X., & Murase, K. (2012). MWMOTE--majority weighted minority oversampling technique for imbalanced data set learning. IEEE Transactions on Knowledge and Data Engineering, 26(2), 405-425.

Bhowan, U., Johnston, M., & Zhang, M. (2011). Developing new fitness functions in genetic programming for classification with unbalanced data. IEEE Transactions on Systems, Man and Cybernetics, Part B (Cybernetics), 42(2), 406-421.

Bishop, C. M. (1995). Neural networks for pattern recognition. Oxford university press.

Breiman, L. (1996). Bagging predictors. Machine learning, 24(2), 123-140.

Catania, C. A., Bromberg, F., & Garino, C. G. (2012). An autonomous labeling approach to support vector machines algorithms for network traffic anomaly detection. Expert Systems with Applications, 39(2), 1822-1829.

Charte, F., Rivera, A. J., Del Jesus, M. J., & Herrera, F. (2019). REMEDIAL-HwR: Tackling multilabel imbalance through label decoupling and data resampling hybridization. Neurocomputing, 326, 110-122.

Chawla, N. V., Lazarevic, A., Hall, L. O., & Bowyer, K. W. (2003, September). SMOTEBoost: Improving prediction of the minority class in boosting. In European conference on principles of data mining and knowledge discovery (pp. 107-119). Springer, Berlin, Heidelberg.

Durgesh, K. S., & Lekha, B. (2010). Data classification using support vector machine. Journal of theoretical and applied information technology, 12(1), 1-7.

Ebenuwa, S. H., Sharif, M. S., Alazab, M., & Al-Nemrat, A. (2019). Variance ranking attributes selection techniques for binary classification problem in imbalance data. IEEE Access, 7, 24649-24666.

Eggermont, J., Kok, J. N., & Kosters, W. A. (2004, March). Genetic programming for data classification: Partitioning the search space. In Proceedings of the 2004 ACM symposium on Applied computing (pp. 1001-1005).

Elhassan, T., & Aljurf, M. (2016). Classification of imbalance data using tomek link (t-link) combined with random under-sampling (rus) as a data reduction method.

Elreedy, D., & Atiya, A. F. (2019, June). A novel distribution analysis for smote oversampling method in handling class imbalance. In International Conference on Computational Science (pp. 236-248). Springer, Cham.

Farid, D. M., Zhang, L., Hossain, A., Rahman, C. M., Strachan, R., Sexton, G., & Dahal, K. (2013). An adaptive ensemble classifier for mining concept drifting data streams. Expert Systems with Applications, 40(15), 5895-5906.

Fernández, A., del Río, S., Chawla, N. V., & Herrera, F. (2017). An insight into imbalanced big data classification: outcomes and challenges. Complex & Intelligent Systems, 3(2), 105-120.

Friedl, M. A., & Brodley, C. E. (1997). Decision tree classification of land cover from remotely sensed data. Remote sensing of environment, 61(3), 399-409.

Gong, J., & Kim, H. (2017). RHSBoost: Improving classification performance in imbalance data. Computational Statistics & Data Analysis, 111, 1-13.

Haykin, S. (1999). Self-organizing maps. Neural networks-A comprehensive foundation, 2nd edition, Prentice-Hall.

He, H., Bai, Y., Garcia, E. A., & Li, S. A. (2008). Adaptive synthetic sampling approach for imbalanced learning. IEEE International Joint Conference on Neural Networks. 2008.

Hu, Y., Guo, D., Fan, Z., Dong, C., Huang, Q., Xie, S., ... & Xie, Q. (2015). An improved algorithm for imbalanced data and small sample size classification. Journal of Data Analysis and Information Processing, 3(03), 27.

Jedrzejowicz, J., Kostrzewski, R., Neumann, J., & Zakrzewska, M. (2018). Imbalanced data classification using MapReduce and relief. Journal of Information and Telecommunication, 2(2), 217-230.

Junsomboon, N., & Phienthrakul, T. (2017, February). Combining over-sampling and under-sampling techniques for imbalance dataset. In Proceedings of the 9th International Conference on Machine Learning and Computing (pp. 243-247).

Kaur, P., & Gosain, A. (2018). Comparing the behavior of oversampling and undersampling approach of class imbalance learning by combining class imbalance problem with noise. In ICT Based Innovations (pp. 23-30). Springer, Singapore.

Kotu, V., & Deshpande, B. (2018). Data science: concepts and practice. Morgan Kaufmann.

Koziarski, M., Kwolek, B., & Cyganek, B. (2018). Convolutional neural network-based classification of histopathological images affected by data imbalance. In Video Analytics. Face and Facial Expression Recognition (pp. 1-11). Springer, Cham.

Lango, M., & Stefanowski, J. (2018). Multi-class and feature selection extensions of roughly balanced bagging for imbalanced data. Journal of Intelligent Information Systems, 50(1), 97-127.

Leevy, J. L., Khoshgoftaar, T. M., Bauder, R. A., & Seliya, N. (2018). A survey on addressing high-class imbalance in big data. Journal of Big Data, 5(1), 42.

Machová, K., Puszta, M., Barčák, F., & Bednár, P. (2006). A comparison of the bagging and the boosting methods using the decision trees classifiers. Computer Science and Information Systems, 3(2), 57-72.

Maldonado, S., Weber, R., & Famili, F. (2014). Feature selection for high-dimensional class-imbalanced data sets using Support Vector Machines. Information sciences, 286, 228-246.

Mitchell, T. M. (1997). Machine Learning, McGraw-Hill Higher Education. New York.

Popel, M. H., Hasib, K. M., Habib, S. A., & Shah, F. M. (2018, December). A Hybrid Under-Sampling Method (HUSBoost) to Classify Imbalanced Data. In 2018 21st International Conference of Computer and Information Technology (ICCIT) (pp. 1-7). IEEE.

Rahman, C. M., Farid, D. M., & Rahman, M. Z. (2011). Adaptive intrusion detection based on boosting and naive bayesian classifier.

Rout, N., Mishra, D., & Mallick, M. K. (2018). Handling imbalanced data: A survey. In International Proceedings on Advances in Soft Computing, Intelligent Systems and Applications (pp. 431-443). Springer, Singapore.

Sanabila, H. R., & Jatmiko, W. (2018, May). Ensemble learning on large scale financial imbalanced data. In 2018 International Workshop on Big Data and Information Security (IWBIS) (pp. 93-98). IEEE.

Schapire, R. E. (1990). The strength of weak learnability. Machine learning, 5(2), 197-227.

Seiffert, C., Khoshgoftaar, T. M., Van Hulse, J., & Napolitano, A. (2009). RUSBoost: A hybrid approach to alleviating class imbalance. IEEE Transactions on Systems, Man and Cybernetics-Part A: Systems and Humans, 40(1), 185-197.

Showrov, M. I. H., & Sobhan, M. (2019, September). Keyword Extraction from Bengali News. In 2019 5th International Conference on Advances in Electrical Engineering (ICAEE) (pp. 658-662). IEEE.

Showrov, M. I. H., Al Awal, M. A., & Sazzad, S. (2019a, April). Identification of Users Feature Based on Facebook Snippets. In 2019 International Conference on Advances in Computing and Communication Engineering (ICACCE) (pp. 1-5). IEEE.

Showrov, M. I. H., Islam, M. T., Hossain, M. D., & Ahmed, M. S. (2019b, December). Performance Comparison of Three Classifiers for the Classification of Breast Cancer Dataset. In 2019 4th International Conference on Electrical Information and Communication Technology (EICT) (pp. 1-5). IEEE.

Singh, A., & Purohit, A. (2015). A survey on methods for solving data imbalance problem for classification. International Journal of Computer Applications, 127(15), 37-41.

Sun, Z., Song, Q., Zhu, X., Sun, H., Xu, B., & Zhou, Y. (2015). A novel ensemble method for classifying imbalanced data. Pattern Recognition, 48(5), 1623-1637.

Verbiest, N., Ramentol, E., Cornelis, C., & Herrera, F. (2012, November). Improving SMOTE with fuzzy rough prototype selection to detect noise in imbalanced classification data. In Ibero-American Conference on Artificial Intelligence (pp. 169-178). Springer, Berlin, Heidelberg.

Wang, Y. (2019, August). An Ensemble Learning Imbalanced Data Classification Method Based on Sample Combination Optimization. In Journal of Physics: Conference Series (Vol. 1284, No. 1, p. 012035). IOP Publishing.

Wei, X., Jiang, F., Wei, F., Zhang, J., Liao, W., & Cheng, S. (2017, May). An ensemble model for diabetes diagnosis in large-scale and imbalanced dataset. In Proceedings of the computing frontiers conference (pp. 71-78).

Yijing, L., Haixiang, G., Xiao, L., Yanan, L., & Jinling, L. (2016). Adapted ensemble classification algorithm based on multiple classifier system and feature selection for classifying multi-class imbalanced data. Knowledge-Based Systems, 94, 88-104.

Yu, H., & Ni, J. (2014). An improved ensemble learning method for classifying high-dimensional and imbalanced biomedicine data. IEEE/ACM transactions on computational biology and bioinformatics, 11(4), 657-666.

Yu, H., Sun, C., Yang, X., Yang, W., Shen, J., & Qi, Y. (2016). ODOC-ELM: Optimal decision outputs compensation-based extreme learning machine for classifying imbalanced data. Knowledge-Based Systems, 92, 55-70.